\documentclass[letterpaper, 10 pt, conference]{ieeeconf}  

\IEEEoverridecommandlockouts                              

\usepackage{graphics} 
\usepackage{epsfig} 
\usepackage{amssymb}  
\usepackage{algpseudocode}
\usepackage{algorithm}
\usepackage{algpseudocode}
\usepackage[dvipsnames]{xcolor}
\usepackage{threeparttable}

\title{\LARGE \bf
State Dropout-Based Curriculum Reinforcement Learning for Self-Driving at Unsignalized Intersections*
}

\author{Shivesh Khaitan$^{1}$ and John M. Dolan$^{2}$
\thanks{*This work was supported by the CMU Argo AI Center for Autonomous Vehicle Research.}
\thanks{$^{1, 2}$The authors are with the Robotics
Institute, Carnegie Mellon University, Pittsburgh, PA, the USA {\tt\small \{shiveshk,jdolan\}@andrew.cmu.edu}}%
}

\begin{document}

\maketitle
\thispagestyle{empty}
\pagestyle{empty}

\begin{abstract}

Traversing intersections is a challenging problem for autonomous vehicles, especially when the intersections do not have traffic control. Recently deep reinforcement learning has received massive attention due to its success in dealing with autonomous driving tasks. In this work, we address the problem of traversing unsignalized intersections using a novel curriculum for deep reinforcement learning. The proposed curriculum leads to: 1) A faster training process for the reinforcement learning agent, and 2) Better performance compared to an agent trained without curriculum. Our main contribution is two-fold: 1) Presenting a unique curriculum for training deep reinforcement learning agents, and \textcolor{black}{2) showing the application of the proposed curriculum for the unsignalized intersection traversal task. The framework expects processed observations of the surroundings from the perception system of the autonomous vehicle. We test our method in the CommonRoad motion planning simulator on T-intersections and four-way intersections.}

\end{abstract}

\section{INTRODUCTION}

Urban intersections pose a difficult driving challenge. Even for human drivers, traversing an intersection requires undivided attention. The situation gets much more complicated without traffic signals. Unsignalized intersections require much more coordination among the incoming vehicles. \textcolor{black}{Such intersections have seen many fatal accidents. The US Federal Highway Administration has reported that about 71.4\% of intersection-related fatal crashes occur at unsignalized intersections \cite{administration_2022}. Autonomous vehicles can reduce the chances of accidents by eliminating human error. }However, intersections are difficult for autonomous vehicles as well. This is especially because predicting the intent of other incoming drivers from different directions at intersections is in itself a complicated problem for self-driving. 

A popular choice for decision-making and planning at intersections for autonomous vehicles has been rule-based methods \cite{lee1976theory}. These were especially very popular for the DARPA Urban Challenge \cite{urmson2008autonomous, montemerlo2008junior}. Although they make the system interpretable, tuning the parameters to avoid collision in every situation is very hard. The rule-based methods also usually try to be overly cautious, leading to suboptimal behaviors. Moreover, a rule-based system needs to consider the trajectory generator's ability to generate feasible trajectories based on the goal or intent set by the decision-making module, which might not be possible in all circumstances.

Reinforcement learning (RL) and deep reinforcement learning (DRL) have recently gained a lot of attention in the literature for autonomous driving. RL's ability to deal with complex scenarios has been unmatched by rule-based methods. It also does not require massive datasets as supervised learning does and can replace individual software modules or be used as a complete end-to-end solution for driving. Its abilities and performance have been demonstrated in many previous works \cite{9294407, isele2018navigating}. However, complex tasks like autonomous driving usually require long training periods to produce acceptable results. This is because of the high-dimensional state space of the autonomous driving task.

To accelerate the training process, curriculum learning \cite{bengio2009curriculum} has been proposed. The basic idea in curriculum learning is to design the training process as a set of graduated steps with increasing complexity of tasks. It draws inspiration from the way humans learn by starting with easier tasks and gradually increasing the complexity to become an expert. The idea of curriculum learning has also been applied to reinforcement learning. For a detailed review on the applications of curriculum for reinforcement learning, see \cite{narvekar2020curriculum}. Apart from accelerating the training process, it has also been proposed that using a curriculum can help find better local minima as compared to training without curriculum \cite{bengio2009curriculum}.

Previously, most works have used hand-designed task-based curricula for training, which rely on segregating the tasks at hand based on their difficulty and training sequentially with increasing levels of difficulty. \textcolor{black}{Such manual segregation of tasks can be laborious and time-consuming.} In this work, we propose a unique automated curriculum for training that is easy to design and is also more effective in dealing with unsignalized intersections. We test our method using CommonRoad-RL \cite{itsc2021.pdf} on \textcolor{black}{T-intersections} and four-way intersections without traffic control \textcolor{black}{and compare it with standard PPO and rule-based TTC method}. The main contributions of this work are:
\begin{itemize}
    \item An automated curriculum for training deep reinforcement learning agents.
    \item An application of the proposed method to the task of traversing unsignalized intersections.
\end{itemize}

The rest of this paper is organized as follows. Section \ref{sec:related_work} provides a review of some important related work. Section \ref{sec:problem_definition} gives an introduction to the intersection problem addressed in this work. Section \ref{sec:methodology} contains necessary background on key methodologies. Section \ref{sec:experiments} presents the experimental results. The conclusions and the future work are in Section \ref{sec:conclusion}.

\section{RELATED WORK}

\label{sec:related_work}

This section summarizes relevant previous work categorized as follows: a) work related to planning for autonomous vehicles focusing on reinforcement learning and b) techniques to accelerating training.

\subsection{Planning for autonomous vehicles}

In the existing literature, planning for autonomous vehicles has used several methods, including rule-based methods, imitation learning, reinforcement learning, and inverse reinforcement learning.

The most popular rule-based method is the time-to-collision (TTC) method \cite{lee1976theory}. This method relies on hand-crafted rules and parameters to control a vehicle and avoid collisions. Trajectory optimization-based methods have also been extensively used for motion planning. Model Predictive Control (MPC) \cite{mmpc, borrelli2005mpc}, which is a receding horizon framework with repeated optimization, is a widely used method. \cite{pan2020} proposed the use of a prediction framework combined with constrained iterative LQR, which is a lightweight optimization method as compared to MPC. However, designing the heuristics for optimization is extremely laborious and often leads to suboptimal results, as they do not adapt well to different driving environments.

Recently, learning-based methods have been quite successful for many robotics and autonomous systems tasks. For self-driving, imitation learning leverages expert driving data to train an autonomous driver, which can work in an end-to-end framework by directly mapping sensor input to trajectory or control commands. Behavior cloning \cite{codevilla2018end} uses supervised learning on collected expert data. However, it suffers from the problem of distribution shift between training and test environments. This is tackled by online interactive data collection methods \cite{ross2011reduction, 8968287} for imitation learning, which require an online demonstrator at training time. 
These generalize much better to varied scenarios, but the requirement of an expert demonstrator itself makes training challenging.

Deep reinforcement learning is a promising framework for complex self-driving planning tasks. Recently, it has been extensively used for motion planning in complex environments \cite{tai2017virtual, everett2018motion}. However, certain complex driving scenarios like unsignalized intersections still remain a challenge. \cite{tram2019learning} uses a hierarchical framework with a high-level RL-based decision-making planner and a low-level MPC controller. However, it does not consider cases when the ego-vehicle might have to turn left or right at intersections instead of just going straight. \cite{7353877} deals with this uncertainty at intersections, but it focuses on T-intersections only, which is a simpler scenario as compared to four-way intersections. \cite{8593420} used Deep Q-learning Network (DQN) for intersection handling. However, DQNs can only be used for problems with a discrete action space and hence cannot be used for continuous control commands. The Deep Deterministic Policy Gradient (DDPG) method \cite{lillicrap2015continuous} works for continuous control spaces. However, both DQN and DDPG require long training periods to converge. In this work, we use Proximal Policy Optimization (PPO) \cite{schulman2017proximal}, which is a variant of the actor-critic family and uses an adaptive KL divergence penalty to control the change of policy at each iteration. To accelerate training and improve performance, we use the state dropout-based curriculum described in section \ref{sec:curriculum}.

\subsection{Techniques for accelerating training periods}

A major drawback of RL methods is long training periods and convergence to sub-optimal policies. This usually happens because the dynamics of the RL training environments are unknown to the agent, and it takes a sufficient number of interactions with the environment before the agent can understand the dynamics to improve its policy. Sparse rewards and partial observability aggravate the problem. This prohibits the usage of RL in safety-critical systems, as the systems might be required to enter unsafe states while the agent trains. Some techniques to alleviate this problem have been proposed in the literature.

Transfer learning (TL) is a technique which pretrains a network on a different task before training on the target task. The network benefits by using the skills acquired during pretraining. For a comprehensive literature review on transfer learning for RL, see \cite{zhu2020transfer}. Apart from just having the advantage of faster training time, using transfer learning also leads to better generalization and better performance than an agent trained without using the pretrained network.

Curriculum learning \cite{10.1145/1553374.1553380}, another approach to expedite training, uses a specially designed schedule for training. It trains on easier tasks to start with and gradually increases the difficulty level. This draws inspiration from how humans learn to do complex tasks by ``starting small'' \cite{elman1993learning}. The main idea is that once the agent has learned to do a simple task well, it will take fewer iterations to adapt to a harder task having the simpler task as a subset because of the stability of the training process. Apart from just accelerating the training process, it has also been shown that certain tasks have been nearly impossible to train without following a curriculum \cite{openai2019solving}. \cite{song2021autonomous} used curriculum reinforcement learning for overtaking in an autonomous racing scenario. However, hand segregating tasks for designing a curriculum might be cumbersome, as deciding the difficulty level of a task might not be trivial. \cite{8500603} proposes an automatic curriculum generator that tries to alleviate the problem. 

In this work, we propose unique automated curricula for autonomous driving at unsignalized intersections. The curricula are easier to train and reduce the training time. For the current work, we assume perception to be solved so that the framework accurately observes the processed road boundary and obstacle information. However, the curricula can be extended to end-to-end frameworks as well.

\section{PROBLEM DEFINITION} \label{sec:problem_definition}

\subsection{Problem Statement}
The problem at hand is generating control commands for an ego-vehicle to traverse unsignalized intersections and reach a predefined goal state. \textcolor{black}{We consider two types of intersection scenarios, as shown in Fig. \ref{fig:problem_formulation_T} and Fig. \ref{fig:problem_formulation}. The ego-vehicle is depicted in green, target vehicles are in blue and the goal region is in yellow. The dotted lines starting from target vehicles show their future trajectories, which do not change in response to the ego-vehicle behavior. This increases the difficulty level of the problem significantly.}

\subsubsection{T-intersection}
\begin{figure}[thpb]
  \centering
  \includegraphics[width=180pt]{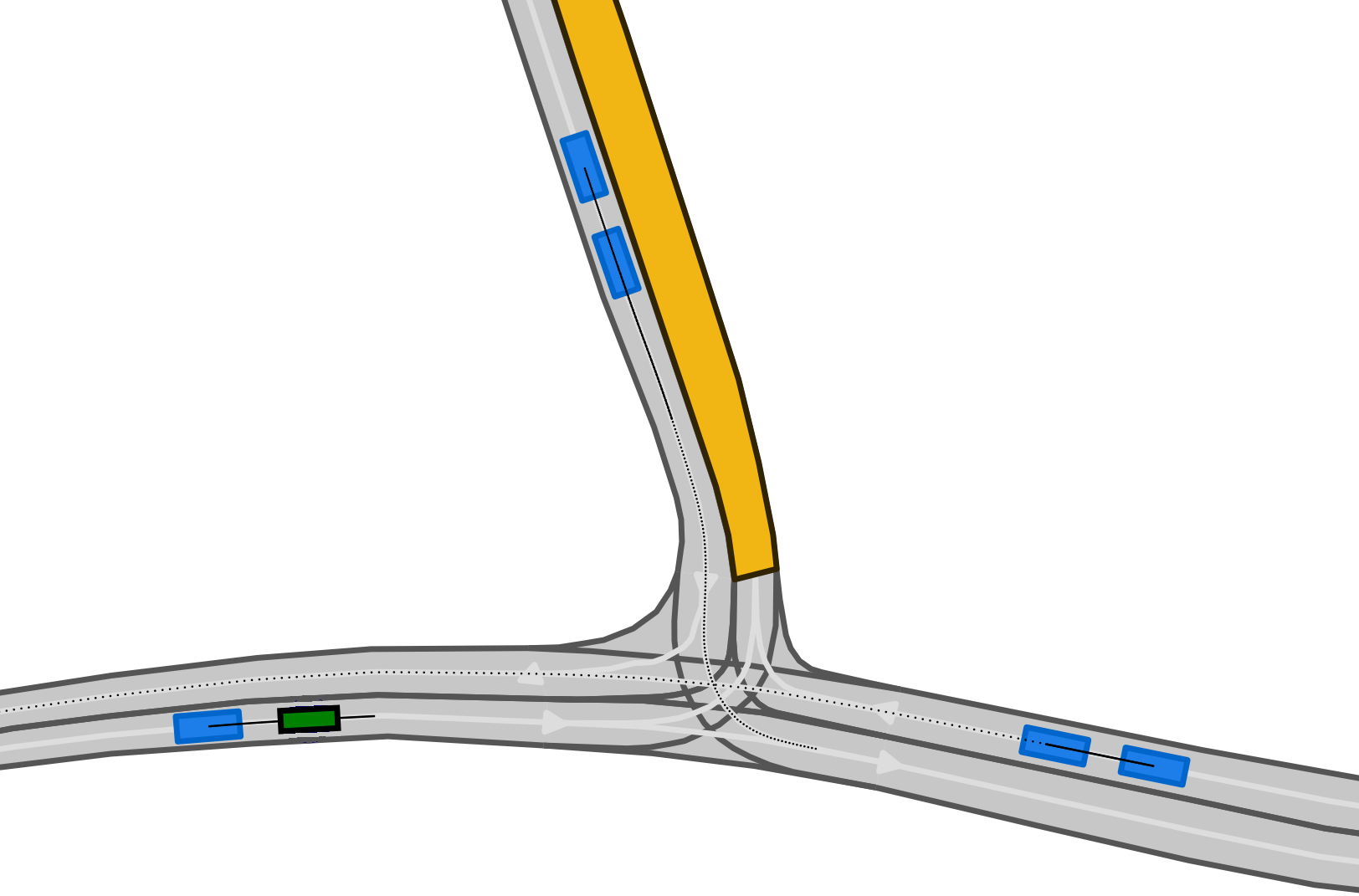}
  \caption{T-intersection}
  \label{fig:problem_formulation_T}
\end{figure}

\textcolor{black}{These scenarios are adopted from the CommonRoad benchmarks: {\tt{ZAM\_Tjunction}} (Fig. \ref{fig:problem_formulation_T}).}
\vspace{-5pt}

\subsubsection{Four-way intersection}

\begin{figure}[thpb]
  \centering
  \includegraphics[width=220pt]{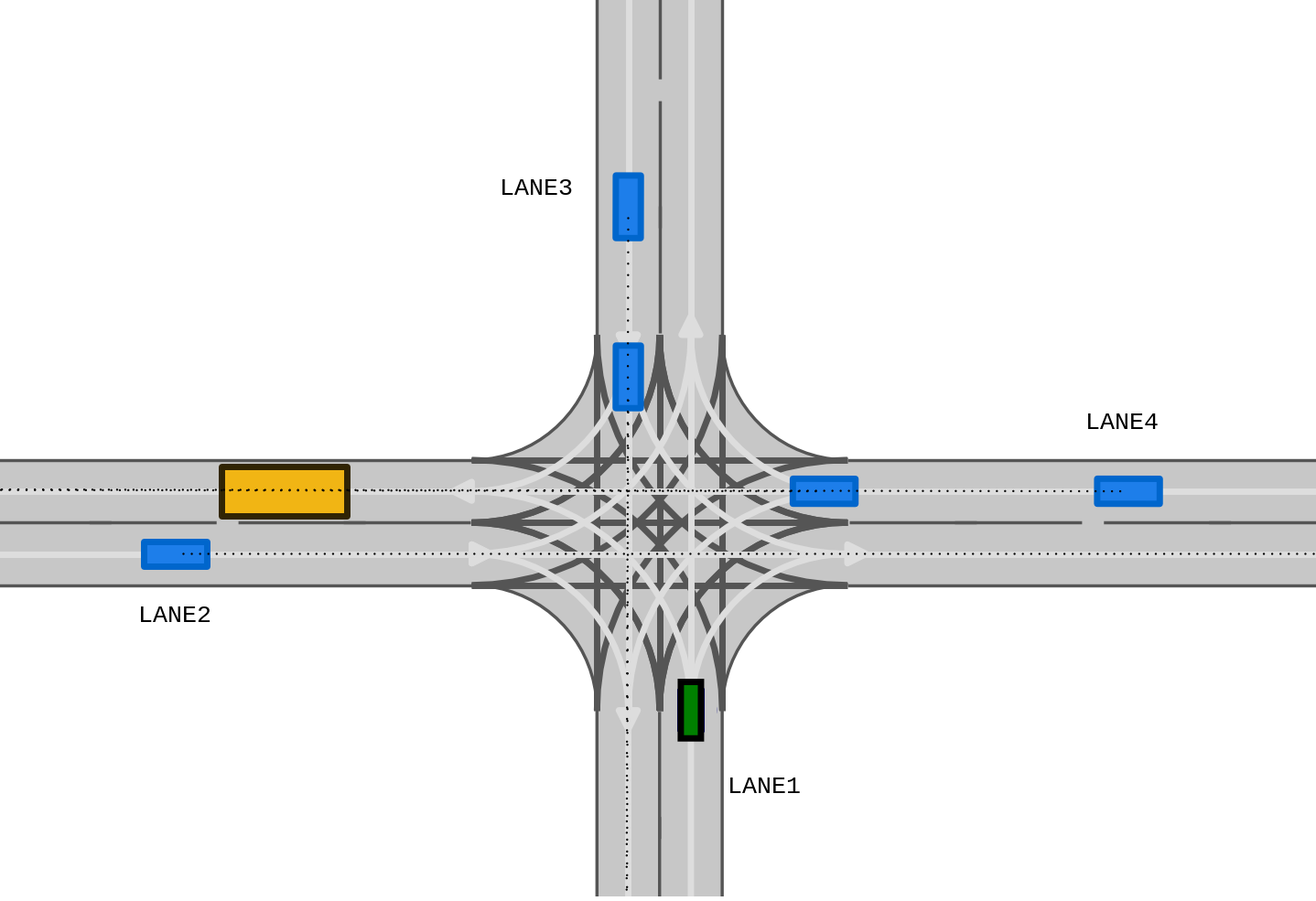}
  \caption{Four-way intersection}
  \label{fig:problem_formulation}
\end{figure}

\textcolor{black}{
These are four-way intersection scenarios (Fig. \ref{fig:problem_formulation}) custom-generated using the CommonRoad scenario designer tool \cite{designer}, as the official CommonRoad benchmarks lacked four-way intersections. The goal state for these scenarios can be in the left, straight-ahead, or right lane. The ego-vehicle has to reach the goal region within $\pm 11.45^\circ$ of the road lane orientation to successfully complete a scenario. The limits on orientation ensure that the ego-vehicle does not reach the goal with an arbitrary orientation, which can make it harder for the vehicle to correct the orientation when moving further in the goal lane. The scenario has to be completed before a time-out of 150 sec.}

\vspace{-10pt}
\vspace{-10pt}

\subsection{Observation and Action space}
\label{sec:observation_and_action_space}
The observation space is as defined in Table \ref{tab:observation_space}. The observations can be divided into four categories: 1) ego-vehicle state; 2) goal and reference path-related observations; 3) surrounding vehicles-related observations; and 4) road network-related observations. The dynamics constraints are based on the kinematic bicycle model ({\tt{KS1}}) as described in \cite{Althoff2017a}. The reference path observation consists of waypoints from the path generated by A$^*$ search over the lanelet network in Commonroad. The distance advancements for longitudinal and lateral directions are calculated along this reference path. For the surrounding vehicles, we include the current and future states (for 5 time-steps with a time discretization of 0.1 sec) of each of the five vehicles in the ego-vehicle frame. The information in future states is as described in section \ref{sec:curriculum}.

The action space consists of acceleration and steering for the ego-vehicle $A = \{accel, steer\} $. The controls are continuous.

\vspace{-5pt}
\begin{table}[ht]
\setlength{\tabcolsep}{3pt} 
\begin{threeparttable}
\caption{Observation Space}
\begin{center}
\begin{tabular}{|c|c|c|}
\hline
& \textbf{Description} & \textbf{Values}\\
\hline\hline
$v_t$ & Ego-vehicle absolute velocity & $\mathbb{R}$\\
\hline
$a_{t-1}$ & Ego-vehicle previous acceleration & $\mathbb{R}$\\
\hline
$\delta_{t-1}$ & Ego-vehicle previous steering & $\mathbb{R}$\\
\hline
$\theta_t$ & Ego-vehicle orientation & $\mathbb{R}$\\
\hline\hline
$r$ & Reference path waypoints ($x,y,\theta$) in ego-vehicle frame & $\mathbb{R}^{15}$\\
\hline
$d\theta_t$ & Ego-vehicle orientation deviation from reference path & $\mathbb{R}$\\
\hline
$d_{long}$ & Longitudinal distance advancement towards goal & $\mathbb{R}$\\
\hline
$d_{lat}$ & Latitudinal distance advancement towards goal & $\mathbb{R}$\\
\hline
$t_{out}$ & Time remaining before time-out & $\mathbb{R}$\\
\hline\hline
$c$ & Lane curvature & $\mathbb{R}$\\
\hline
$l_o$ & Ego-vehicle offset from lane centerline & $\mathbb{R}$\\
\hline
$r_l$ & Ego-vehicle distance from left road boundary & $\mathbb{R}$\\
\hline
$r_r$ & Ego-vehicle distance from right road boundary & $\mathbb{R}$\\
\hline
$l_l$ & Ego-vehicle distance from left lane boundary & $\mathbb{R}$\\
\hline
$l_r$ & Ego-vehicle distance from right lane boundary & $\mathbb{R}$\\
\hline \hline
$o$ & Target vehicles' current and future states$^*$($x, y, v, \theta$) & $\mathbb{R}^{100}$\\
\hline
\end{tabular}
\begin{tablenotes}
  \item $*$ The information in future states depends on the curriculum
\end{tablenotes}
\end{center}
\end{threeparttable}
\label{tab:observation_space}
\end{table}
\vspace{-8pt}

\section{METHODOLOGY}
\label{sec:methodology}
This section introduces the reward structure for the PPO agent, and the proposed curricula for learning.

\subsection{Reward Structure}
\label{sec:reward_structure}
The components of the reward function are as follows:

\begin{itemize}
    \item A positive reward for distance advancement along the reference path as described in section \ref{sec:observation_and_action_space}. The reward is calculated as $\rho_1 d_{long} + \rho_2 d_{lat}$ where $\rho_1$ and $\rho_2$ are positive constants.
    \item A constant negative reward $\sigma_1$ for collision with obstacles, going off-road, and violating dynamics constraints.
    \item A constant negative reward $\sigma_2$ for time-out.
    \item A constant positive reward $\phi$ for reaching the goal within the allowed orientation error.
\end{itemize}

An episode is terminated when any one of the following conditions is met: ego-vehicle reaches the goal within allowed orientation error, ego-vehicle goes off-road, ego-vehicle collides with another obstacle, applied controls violate dynamics constraints or time-out.

\subsection{State dropout-based curriculum for PPO}
\label{sec:curriculum}

Training and getting an acceptable performance with PPO or other existing RL algorithms in the described intersection problem is tough. Often the RL agents settle for a suboptimal policy. A prominent reason for this is that once an RL agent adopts a suboptimal policy, it further keeps exploiting the suboptimal policy and takes a long time to escape local minima and generate better actions after that. Moreover, unsignalized intersections also require that a driver is able to determine the intent of multiple other incoming vehicles. Training a simple PPO agent in this scenario leads to a suboptimal behavior even with long training periods, as implicitly learning to predict the behavior of the vehicles while simultaneously learning to control the car is not trivial for the PPO agent.

Our state dropout-based curriculum overcomes the challenge by ``starting small''. An easy scenario for an RL agent in intersections is when the intentions of the surrounding drivers are known in advance. In such scenarios, the RL agent can learn an optimal policy for traversing. Thus we incorporate the future state information (for $N$ time-steps) about the surrounding vehicles in the observation space for PPO. We propose two different curricula to deal with the additional privileged observation:

\subsubsection{Curriculum 1}
In this method we train the agent in $N + 1$ phases starting with phase 0. We begin training with the future state information for $N$ time-steps. After phase 0, we drop the $N^{th}$ future state information and continue training. In the subsequent phases we keep on dropping more future state information and in the final phase fine-tune the agent without any future state information. Algorithm \ref{alg:curriculum1} describes the $STEP$ function for this method.

\begin{algorithm}
\caption{Step function for curriculum 1}\label{alg:curriculum1}
\begin{algorithmic}[1]
\Procedure{step}{\textbf{action, phase}}
\State Apply acceleration and steering from \textbf{action}
\State \textbf{obs} = Observe()
\State \textbf{reward}, \textbf{done} = Reward(\textbf{obs})

\For{\texttt{$i \leftarrow (N - phase + 1)$} to $N$}
    \State Drop $i^{th}$ future prediction from \textbf{obs}
\EndFor

\State \textbf{return} obs, reward, done
\EndProcedure
\end{algorithmic}
\end{algorithm}

\subsubsection{Curriculum 2} In this method, instead of dropping the future state information sequentially after each phase, we augment the action space of the agent with an additional action such that the augmented action space is $A = \{accel, steer, pred\}$. Here $pred$ determines which future state information for the surrounding vehicles will be dropped from the observation in the next step. The reward structure as defined in section \ref{sec:reward_structure} is also augmented to add constant positive rewards $\psi_i$ for choosing to drop the future states from the $i^{th}$ to the $N^{th}$ time-step. The decision is based on parameters $\kappa_i$. Here, $\forall i \in [2, N]$ 
$$
 \psi_i < \psi_{i-1} \eqno{(1)}
$$
$$
 \kappa_i < \kappa_{i-1} \eqno{(2)}
$$

This incentivizes the agent to drop more future states. Algorithm \ref{alg:curriculum2} defines the modified $STEP$ function for this method. Thus we let the network choose when to drop the future states. At the beginning of training, when the agent has not learned to predict the behavior of the vehicles, it chooses to use the future information. As the training progresses and the agent learns to control the ego-vehicle, $\psi_i$ incentivizes the agent to learn to predict the driving intention as well in order to get higher rewards. 

An added advantage of this method is that the $pred$ value can be used at test time to determine whether the agent understands the behavior of the other vehicles or not. This can further be used as a confidence score for the agent commands.

\vspace{-5pt}
\begin{algorithm}
\caption{Step function for curriculum 2}\label{alg:curriculum2}
\begin{algorithmic}[1]
\Procedure{step}{\textbf{action}}
\State Apply acceleration and steering from \textbf{action}
\State \textbf{obs} = Observe()
\State \textbf{reward}, \textbf{done} = Reward(\textbf{obs})

\For{\texttt{$i \leftarrow 1$} to $N$}
\If{$action.pred \geq \kappa_i$}
    \State Drop $i^{th}$ future prediction from \textbf{obs}
    \State \textbf{reward} = \textbf{reward} + $\psi_i$
\EndIf
\EndFor

\State \textbf{return} obs, reward, done
\EndProcedure
\end{algorithmic}
\end{algorithm}

\vspace{-12pt}

\section{EXPERIMENTS}
\label{sec:experiments}
The curriculum was tested on the CommonRoad simulator for unsignalized T-intersections and four-way intersections.

\subsection{Experimental Setup}
\textcolor{black}{For the T-intersections, we have 544 {\tt{ZAM\_Tjunction}} scenarios from the CommonRoad benchmarks, which have scenarios with ego-vehicles and target-vehicles initialized at different positions with different velocities. 380 scenarios are used for training, and the rest are used for testing.}

We generated 2000 four-way intersection scenarios. For each scenario, the ego-vehicle is spawned at a randomly sampled location in lane 1 (Fig. \ref{fig:problem_formulation}) with appropriate orientation. The ego-vehicle always starts at rest. The obstacles are spawned at randomly sampled locations in lanes 2, 3 and 4 with appropriate orientation and randomly sampled initial velocities $\in [5, 10]$ m/s. 1400 scenarios were used for training and the rest were used for testing.

\subsection{Network Architecture}
To represent the policies, we used fully-connected (FC) networks with 8 hidden layers of 64 units each and $tanh$ nonlinearities. The hyperparameter values for PPO are listed in table \ref{tab:hyperparameters}. \textcolor{black}{Empirically best performing hyperparameters and network architecture were choosen for the experimental results}. 

\subsection{Training Details}
The training was done on a desktop with 3.5 GHz AMD Ryzen 9 5900hs CPU. We used a customized version of the CommonRoad-RL framework for training and testing. We trained three different agents:

\begin{table}[ht]
\vspace{5pt}
\caption{Hyperparameters}
\begin{center}
\begin{tabular}{|c|c|}
\hline
\textbf{Parameter} & \textbf{Value} \\ \hline
Training Iterations (T-intersections) & 600 \\ \hline
Training Iterations (Four-way intersections) & 650\\ \hline
Discount ($\gamma$) & 0.99 \\ \hline
GAE parameter ($\lambda$) & 0.95 \\ \hline
Clipping parameter & 0.2 \\ \hline
Learning-rate & 0.0005 \\ \hline
No. of environments & 32 \\ \hline
Time horizon & 512 \\ \hline
Batch size & 16384* \\ \hline
\end{tabular}
\end{center}
\begin{center}
\begin{tablenotes}
  \item * Batch size = Time Horizon $\times$ No. of environments
\end{tablenotes}
\end{center}
\label{tab:hyperparameters}
\vspace{-15pt}
\end{table}

\begin{enumerate}

    \item A standard PPO agent without any curriculum. The target vehicles' future state information is dropped by default in the standard agent.
    \item A PPO agent using curriculum 1 with $N = 4$. The agent is thus trained in $5$ phases.
    \item A PPO agent using curriculum 2 with $N = 4$.
\end{enumerate}

\subsection{Performance Evaluation}
To understand the impact of the proposed curricula, we compare the training curves of the agents trained using the curricula with the training curve of the standard PPO agent. We compare the success rate and mean reward to evaluate the performance of our method. \textcolor{black}{The results presented are the best results obtained by repeating each experiment 3 times with a random seed for the network. The future states of the surrounding vehicles are not available during testing for all three methods.}

\subsubsection{T-intersection}
\begin{figure}[thpb]
  \centering
  \includegraphics[width=\columnwidth]{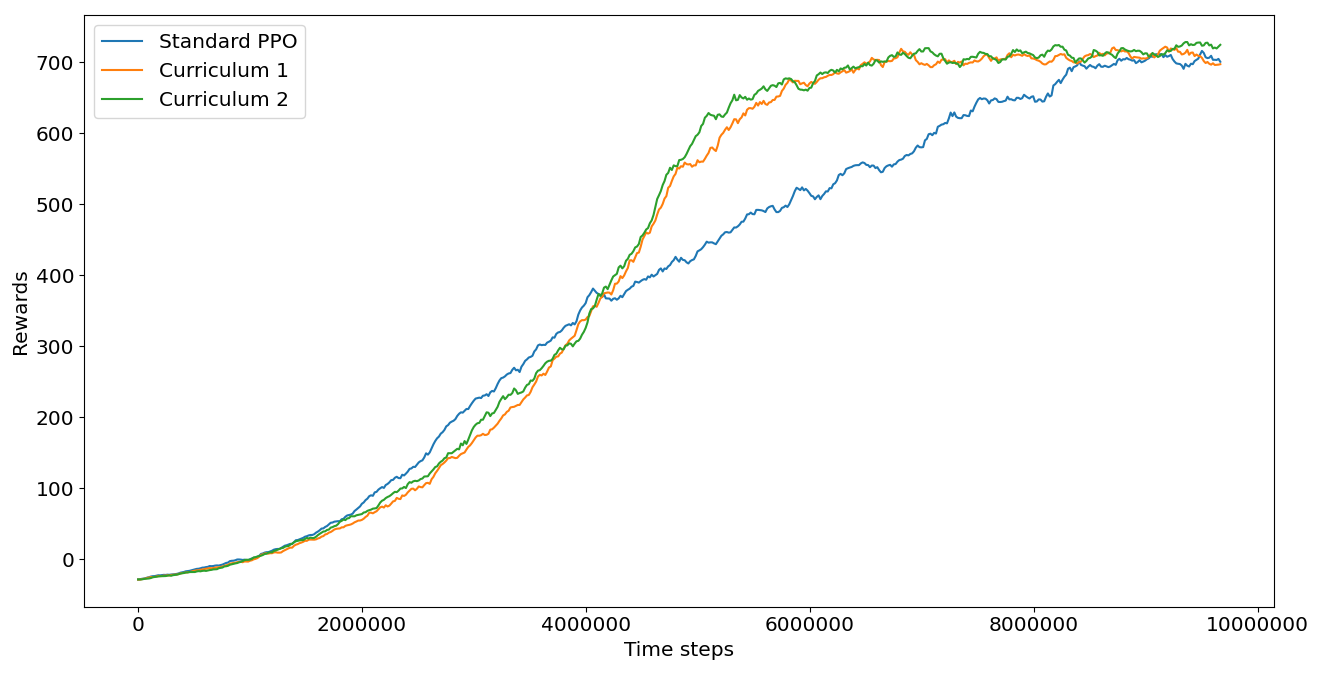}
  \caption{Training curves (T-intersection)}
  \label{fig:training_curves_T}
\end{figure}

\textcolor{black}{
The training curves for the three methods are shown in Fig. \ref{fig:training_curves_T}. The rewards are averaged across 10 updates. It can be seen that both the curriculum-based methods outperform the standard PPO baseline in terms of sample efficiency. While PPO agents with the curricula start converging around 6000000 steps, the baseline standard PPO converges around 8250000 steps.}

\begin{figure}[thpb]
  \centering
  \includegraphics[width=\columnwidth]{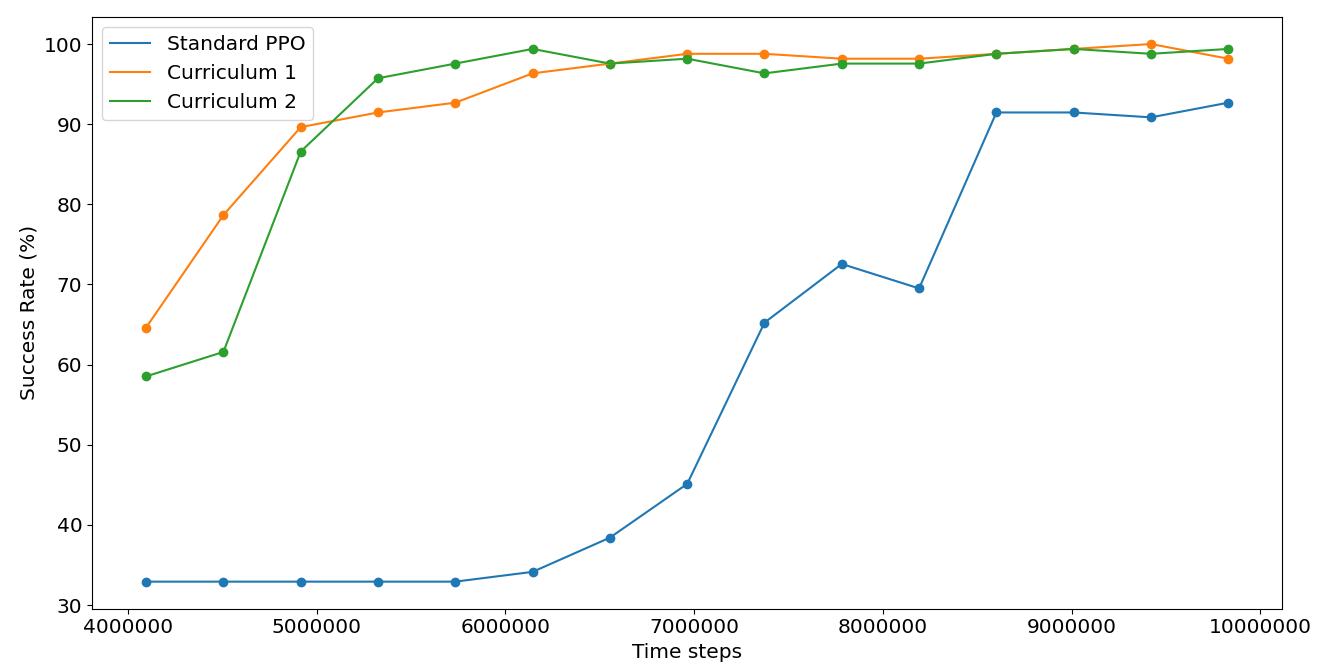}
  \caption{Success rates (T-intersections)}
  \label{fig:success_rates_T}
  \vspace{-5pt}
\end{figure}

\textcolor{black}{Fig. \ref{fig:success_rates_T} shows the success rate of the methods on 164 testing scenarios. Both the curricula outperform the standard PPO baseline with their best performance of 99.39\% as compared to 92.68\% success rate for the baseline.}

\subsubsection{Four-way intersection}
\begin{figure}[thpb]
  \centering
  \includegraphics[width=\columnwidth]{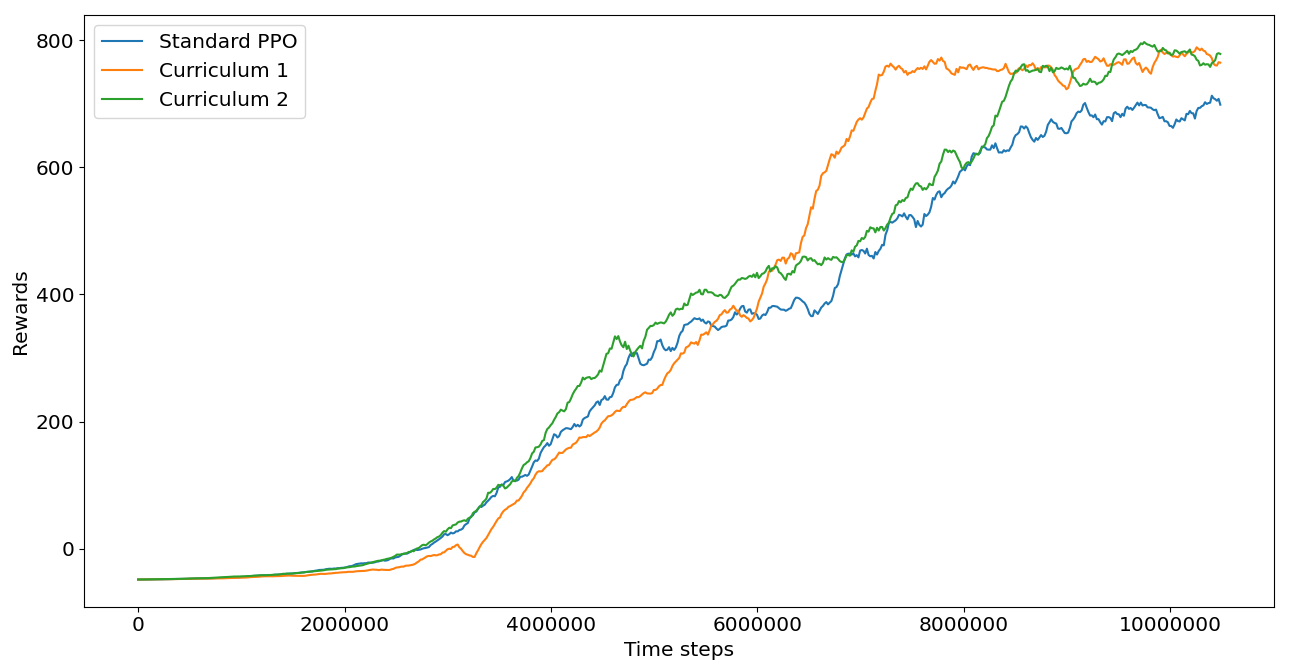}
  \caption{Training curves}
  \label{fig:training_curves}
\end{figure}

\textcolor{black}{
The training curves for the three methods are shown in Fig. \ref{fig:training_curves}. The rewards are averaged across 10 updates. It can be seen that both the curriculum based methods outperform the standard PPO baseline in terms of sample efficiency. While PPO with curriculum 1 starts converging around 7250000 steps, the curriculum 2 approach converges around 8500000 steps. The baseline converges around 9250000 steps. }

\begin{figure}[htp]
\vspace{5pt}
  \centering
  \includegraphics[width=\columnwidth]{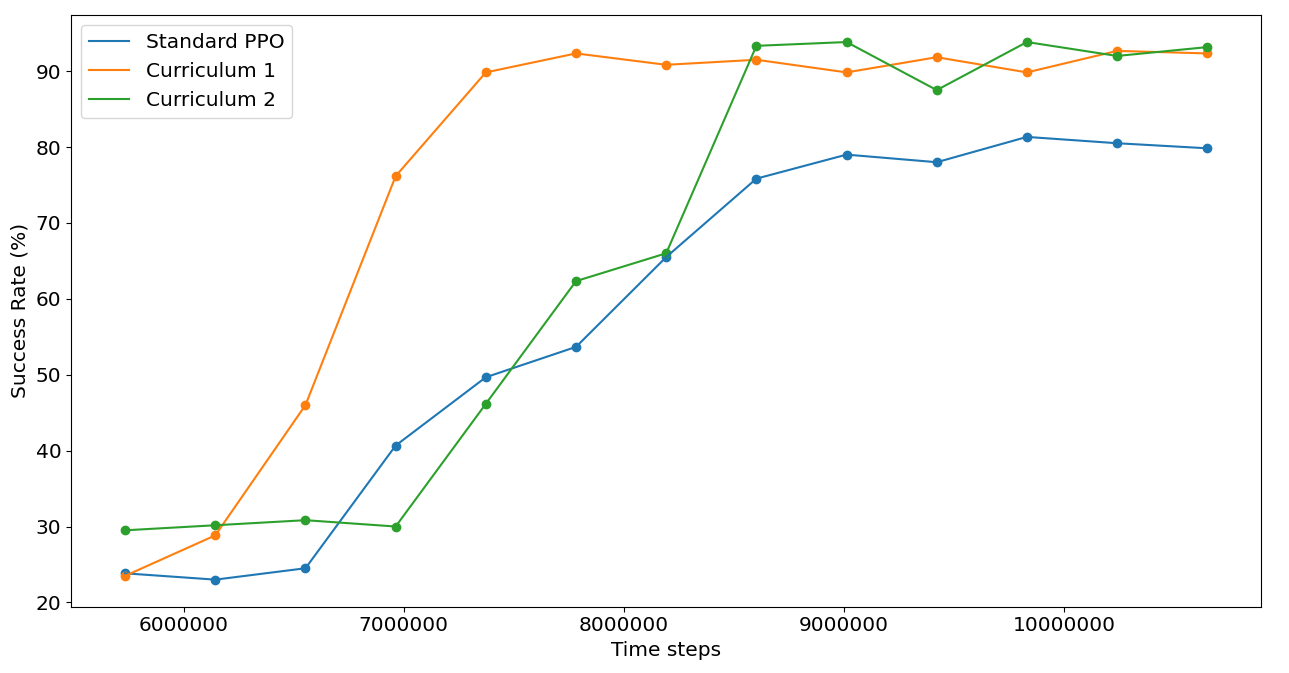}
  \caption{Success rates}
  \label{fig:success_rates}
\end{figure}

Fig. \ref{fig:success_rates} shows the performance of the methods on the 600 testing scenarios. Curriculum 2 shows the best performance with its best model having a success rate of 93.83\%. Curriculum 1 achieves 92.66\% and the baseline achieves 81.33\%.

\textcolor{black}{
We also compare the results with the rule-based TTC method in Table \ref{tab:success_rate}.}

\begin{table}[ht]
\caption{Success rates of curriculum methods versus baselines}
\label{tab:rewards}
\begin{center}
\begin{tabular}{|c|c|c|}
\hline
\textbf{Method} & \textbf{T-intersection (\%)} &  \textbf{Four-way intersection (\%)}\\ \hline
TTC  & 90.85 & 84.33\\
\hline
Standard PPO & 92.68 & 81.33\\ \hline
Curriculum 1 & \textbf{99.39} & 92.66\\ \hline
Curriculum 2 & \textbf{99.39} & \textbf{93.83} \\ \hline
\end{tabular}
\end{center}
\label{tab:success_rate}
\vspace{-15pt}
\end{table}

\subsection{Discussion}
\textcolor{black}{
The training curves and success rates of the curriculum-based agents show that using the curriculum results in higher sample efficiency, with the agents converging much faster as compared to the baseline agent. Also, both the curricula have higher success rates than the baselines in both the scenarios. This shows that the agents converge to better optima as the curriculum makes  the training stable. \textcolor{black}{Though we used ground truth prediction for the target vehicles available in the simulation, for real-world training, predictions from the perception layer can be used. Further avenues of privileged information e.g. bird's eye view in the augmented observation space can also be experimented with.}}

\section{CONCLUSIONS}
\label{sec:conclusion}
\textcolor{black}{
In this work, we propose a novel curriculum for training a deep reinforcement learning agent where future state predictions can achieve faster training and avoid convergence to suboptimal policies. We test the performance of the curriculum on the unsignalized intersection traversal task for autonomous driving. The curriculum outperforms the standard baseline in sample efficiency and test time performance. Though in this paper we show the application of the curriculum to the intersection traversal task only, it has a broader scope and can be generally applied to other similar tasks as well. In future work, we aim to extend the curriculum for end-to-end learning \textcolor{black}{and real-world testing}. }

\addtolength{\textheight}{-12cm}   






\bibliographystyle{IEEEtran}
\bibliography{./IEEEfull,refs}

\end{document}